\documentclass[letterpaper, 10 pt, conference]{ieeeconf}  % Comment this line out if you need a4paper

\IEEEoverridecommandlockouts                              % This command is only needed if % you want to use the \thanks command

\usepackage{cite}
\usepackage{algpseudocode}
\usepackage{algorithm}
\usepackage{graphicx}
\usepackage{cite}
\usepackage{tabularx,booktabs}
\usepackage{amsmath,amsfonts}
\usepackage{array}
\usepackage[caption=false,font=normalsize,labelfont=sf,textfont=sf]{subfig}
\usepackage{textcomp}
\usepackage{stfloats}
\usepackage{url}
\usepackage{verbatim}
\usepackage{textcomp}
\usepackage{xcolor}
\usepackage{url}
\usepackage{hyperref}
\usepackage{siunitx}
\usepackage{caption}
\usepackage{subfig}
\usepackage{multirow}
\title{\LARGE \bf
Identification of Driving Heterogeneity using Action-chains}

\author{Xue Yao*, Simeon C. Calvert and Serge P. Hoogendoorn
\thanks{Xue Yao*, Simeon C. Calvert, and Serge P. Hoogendoorn are with Department of Transport \& Planning, Delft University of Technology, 2628CN. Delft, The Netherlands
        {\tt\small x.yao-3@tudelft.nl, s.c.calvert@tudelft.nl, s.p.hoogendoorn@tudelft.nl}}%
}

\begin{document}

\maketitle

\pagestyle{empty}
\thispagestyle{empty}
%%%%%%%%%%%%%%%%%%%%%%%%%%%%%%%%%%%%%%%%%%%%%%%%%%%%%%%%%%%%%%%%%%%%%%%%%%%%%%%%
\begin{abstract}
Current approaches to identifying driving heterogeneity face challenges in capturing the diversity of driving characteristics and understanding the fundamental patterns from a driving behaviour mechanism standpoint. This study introduces a comprehensive framework for identifying driving heterogeneity from an \emph{Action-chain} perspective. First, a rule-based segmentation technique that considers the physical meanings of driving behaviour is proposed. Next, an \emph{Action phase} Library including descriptions of various driving behaviour patterns is created based on the segmentation findings. The \emph{Action-chain} concept is then introduced by implementing \emph{Action phase} transition probability, followed by a method for evaluating driving heterogeneity. Employing real-world datasets for evaluation, our approach effectively identifies driving heterogeneity for both individual drivers and traffic flow while providing clear interpretations. These insights can aid the development of accurate driving behaviour theory and traffic flow models, ultimately benefiting traffic performance, and potentially leading to aspects such as improved road capacity and safety.
\end{abstract}

%Main body starts
\section{Introduction}
Driving behaviour plays a pivotal role in determining vehicle motion, substantially affecting traffic flow, fuel consumption, and emission. It is widely acknowledged that driving heterogeneity, which is defined as the difference between driving behaviours of driver/vehicle combinations under comparable conditions\cite{ossen2006interdriver}, does exist. 
Research has shown that this heterogeneity contributes to increased traffic accidents and congestion \cite{sun2021modeling}. Additionally, in mixed automated-human traffic, accurate descriptions and predictions of human-driven vehicle (HDV) behaviour are crucial for the decision-making and control of connected and automated vehicles (CAVs). These have underlined the necessity of a better understanding and identification of the heterogeneity in human driving.

It is well established that driving heterogeneity encompasses both intra-heterogeneity, which refers to driver-independent variability, and inter-driving heterogeneity, which involves differences in driving behaviour among drivers \cite{ossen2006interdriver, ossen2011heterogeneity}. However, directly measuring or detecting driving heterogeneity is challenging due to its reliance on human cognitive and physiological processes. With the increasing availability of naturalistic driving data, various efforts have been made to comprehensively and quantitatively analyse driving heterogeneity. The identification of driving heterogeneity from observed driving behaviour is typically approached in two ways \cite{Yao2023}: 1) Employing techniques to characterise driving behaviour by inferring driving profiles from distinct driving events, and 2) Analysing driving behaviour without explicitly creating driving behaviour profiles.

The former approach addressed the identification of driving heterogeneity as a classification or clustering problem, resulting in categorical output with discrete scales or numerical output with continuous scores. For example, Hoogendoorn et al. \cite{hoogendoorn2013driver} developed a method to categorise driver states into low, medium, and high workload categories. Or, clustering techniques have been employed to define a few driving style groups such as aggressive, normal, and mild \cite{sun2021modeling}. However, due to the stochastic and uncertain nature of driving behaviour, these limited groups are insufficient for capturing the diverse characteristics of driving behaviour. Additionally, the criteria used to define these groups are somewhat ambiguous and subjective, posing challenges in effectively eliminating individual biases.

In contrast to employing subjectively defined classes, some research has focused on identifying driving heterogeneity by presenting a driving style space containing a vast array of categories without explicitly establishing driving behaviour profiles. For example, Qi et al. distinguished driving styles based on a space that included over 20 different types \cite{qi2019recognizing}. Another study converted car-following sequences into a comprehensive array of primitive driving patterns, and the distributions of these patterns were then utilised to analyse individual driving styles \cite{wang2018driving}. This approach allows for the recognition of a greater degree of variability in driving behaviour by encompassing various driving characteristics. However, it is essential to acknowledge that this broader categorisation of driving heterogeneity may lead to reduced clarity of the fundamental driving behaviours and a limited understanding of driving heterogeneity. Consequently, further research in this area is necessary to address these challenges. 

To bridge these research gaps, a novel framework is proposed to identify heterogeneity in longitudinal driving behaviour from an \emph{Action-chain} perspective.  An \emph{Action-chain} is defined as a series of \emph{Action phases} over time. The contributions of this research are two-fold: i) A rule-based segmentation technique is presented to divide driving trajectories, considering the clear physical meanings of driving behaviour. ii) The concept of \emph{Action phase} and \emph{Action-chain} are first introduced to interpret driving behaviours, based on which a method for evaluating driving heterogeneity is proposed. The effectiveness of the framework was evaluated using real-world datasets, and the results demonstrate that the proposed methods can effectively identify driving heterogeneity at both individual drivers and traffic flow levels, providing clear interpretations. This approach offers valuable insights into understanding driving behaviour by uncovering underlying heterogeneity, which supports the development of accurate and robust driving behaviour and traffic flow models.

\section{Framework description}

\subsection{Defining Action Phase and Action-chain}
The concept of ``action points'', which refers to specific moments of change in acceleration during driving \cite{knoop2015relation}, serves as the foundation to introduce the concept of \emph{action trend} in this study. While action points capture acceleration or deceleration, they do not fully capture the complexity of driving behaviours. To overcome this limitation, we further propose the concept of \emph{Action phase}, which expands the scope by incorporating additional variables to provide more comprehensive information about driving behaviour.

\begin{figure*}[!t]
\centering
 \includegraphics[width = 0.83\linewidth]{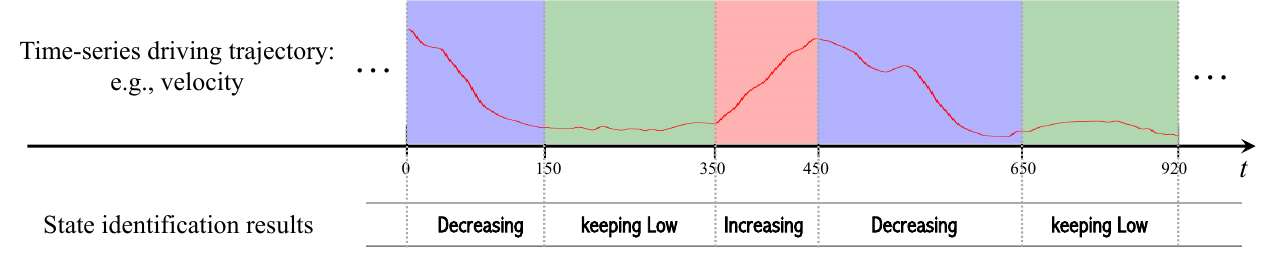}
 \caption{Visualization of time-series driving trajectory: An example of velocity}
 \label{time-series}
\end{figure*}

By examining the univariate trajectory of driving behaviour, illustrated by the example of velocity ($v$) in Figure \ref{time-series}, distinct states are obviously observed. Some trajectories exhibit upward trends, others display downward trends, while some maintain a relatively stable range of fluctuations that can be considered as a keeping trend. We refer to these moments of driving behaviour different tendencies as \emph{action trends}, which are segmented by turning points. Specifically, \emph{action trends} are classified as ``Increasing (I)'', ``Decreasing (D)'', or ``Stable (S)''. To further refine the ``Stable'' trend, it is categorised as ``Stable in a high value (H)'' or ``Stable in a lower value (L)''. Thus, the \emph{action trend} space can be represented as $S = \{I, D, H, L\}$, and the driving trajectory shown in Figure \ref{time-series} can be expressed as $S_v = \{D, L, I, D, L\}$.

It is worth noting that while driving behaviour variables often exhibit synchronisation, our definition of \emph{action trends} allows for variations in the temporal changes of different variables. For example, when the velocity state is ``Increasing'', the acceleration state can be ``Increasing'', ``Decreasing'', ``Stable'', or a combination of them. Consequently, the definition of \emph{action trends} can be extended to other driving behaviour variables, such as acceleration and space headway. Thereafter, the concept of \emph{Action phase} is proposed by encompassing multiple variables, and each \emph{Action phase} label consists of multiple \emph{action trend} names. These \emph{action trend} names are estimated using uniform criteria derived from the group level of drivers in a certain traffic flow.

To account for the inherent sequential nature of driving behaviour, it is essential to consider the temporal dependencies between \emph{Actions phases}. Hence, the concept of an \emph{Action-chain} is introduced to represent a sequence of \emph{Action phases} and their relations. The behaviour of a vehicle over time may consist of one or more \emph{Action-chains}, each corresponding to different responses to the environment. With the \emph{Action-chain} structure, driving behaviour over time can be characterised, which provides valuable insights into the underlying patterns and heterogeneity of driving behaviour. 

\subsection{Introducing the Novel Framework}

\begin{figure*}[!t]
\centering
 \includegraphics[width = 0.92\linewidth]{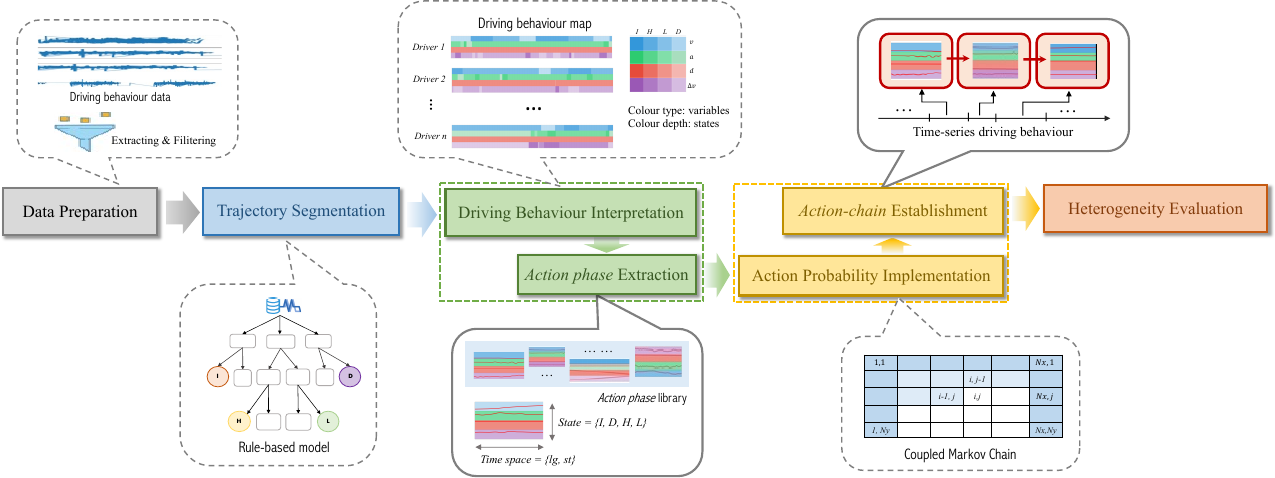}
 \caption{A novel framework of identifying driving heterogeneity}
 \label{framework}
\end{figure*}

The proposed framework for identifying driving heterogeneity aims to estimate frame-wise driving trajectories and identify driving heterogeneity within specific traffic flow conditions. The entire procedure is illustrated in Figure \ref{framework}, consisting of five main steps: Data Preparation, Trajectory Segmentation, Action phase Extraction, Action-chain Establishment, and Heterogeneity Evaluation. The extraction of \emph{Action phase} and the establishment of \emph{Action-chains} involve the preceding steps called Driving Behaviour Interpretation and Action-chain Implementation, respectively.

Data plays a crucial role in the identification of heterogeneity and serves as a fundamental aspect of the analytical process. After data tracing and preprocessing, the time-series driving behaviour data are used as input for the segmentation algorithm (\textbf{Algorithm 1}). It is represented as $x_1, x_2, ..., x_t$, where $x_t$ denotes the driving behaviour variable feature at the $t$-th frame. The segmentation algorithm (\textbf{Algorithm 1}) outputs $l^m_n$, which represents the \emph{action trend} names of variable $m$ to be recognised for the $n$-th segment, where $n = 1, 2, ..., N$. Based on the segmentation results, the driving behaviour of individual drivers can be visualised using driving behaviour maps, which highlight the unique characteristics of each driver. In the driving behaviour map, at the $t$-th frame, the state of driving behaviour is denoted as $S_t = \{l^1, l^2, ..., l^m\}$. Subsequently, \textbf{Algorithm 2} is designed to detect driving behaviour segments in which all variables have a single \emph{action trend}. The output, denoted as \emph{Action phase} and represented as $S_{n'} = \{l^1, l^2, ..., l^m\}$, signifies the \emph{Action phase} for the $n'$-th segment, where $n' \in N'$, $N'$ denotes the total number of \emph{Action phases} for an individual driver. All the output \emph{Action phases} form the \emph{Action phase} Library under a specific traffic flow. The actual size of this Library is generally smaller than the theoretical value $m^4$ due to the nonexistence of certain state combinations in the real world, in accordance with fundamental driving behaviour theories. The length of the \emph{Action phase} at the $n'$-th segment, referred to as the time label, is denoted as $\mathcal{T}_{n'}$.

Considering the time-series nature of driving behaviour, an \emph{Action phase} transition probability algorithm (\textbf{Algorithm 3}) is implemented to capture the temporal dependencies between \emph{Action phases}. An \emph{Action phase} and the next \emph{Action phase} obtained through the maximum transition probability constitute an \emph{Action-chain}, representing the most probable driving behaviour adopted by drivers. The \emph{Action-chain} serves as a description of homogeneous driving behaviour and is used to distinguish heterogeneity in driving behaviour. Drivers who deviate more from the \emph{Action-chains} are considered to exhibit greater heterogeneity (conducted by \textbf{Algorithm 4}).

\section{Method implementation}

\subsection{Rule-based Segmentation}
Traditional classification algorithms, such as the K-nearest neighbour method, support vector machines, and Convolutional Neural Networks have been commonly used for classifying driving styles or recognising driving patterns \cite{Yao2023peformance}. However, the segments obtained using these algorithms often lack clear interpretability in terms of physical characteristics. In contrast, rule-based segmentation is a relatively simple and interpretable method for dividing driving behaviour trajectories into meaningful segments. Therefore, we propose a rule-based method, referred to as \textbf{Algorithm 1} within the framework, to segment driving behaviour trajectories.

Let $V = \{v_1, v_2, ..., v_m\}$ be a set of driving behaviour variables, such as velocity, acceleration, distance, etc. $P = \{(x_1, y_1), ..., (x_n, y_n)\}$ represents a set of turning points for a single variable, which are calculated using calculus, specifically the first and second derivatives. \textbf{Algorithm 1} consists of the following steps:

1. Data preparation: Load the turning points of the selected variable. Calculate the variable changes $\Delta y$ and time intervals $\Delta x$ between neighbouring turning points.

2. Threshold setting: Define threshold values $\theta_1, \theta_2$ to differentiate between segments with state Increasing (I), Decreasing (D), or Stable (S). Set $\gamma$ to determine whether a segment is too short and should be merged with its neighbouring segments.

3. Initial categorisation: If $\Delta y > \theta_1$, meaning that the variable increases to a certain extent, which cannot be ignored, then label the segment as I. If $\Delta y < \theta_2$, in which case the variable decreases to a non-negligible level, then label it as D. When $\theta_2 < \Delta y < \theta_1$, the variable keeps within a small range of changes and is labelled as S.

4. Merging: For each segment labelled as S, if the time interval $\Delta x_{n} < \gamma$, and $\Delta x_{n-1} > \gamma, \Delta x_{n+1} > \gamma$, merge the segment with its neighbouring segment $n+1$.

5. ``Stable'' refinement: For the updated S segments, calculate the mean value of the variable for each segment. Update the labels S as stable in High (H) or Low (L) values based on the threshold $\delta$.

By implementing the above \textbf{Algorithm 1}, each variable in $V$ is assigned \emph{action trend} labels with clear physical meanings. This rule-based method allows for effective segmentation of driving behaviour trajectories based on single variables.

Subsequently, \textbf{Algorithm 2} is proposed to extract \emph{Action phases} with simple steps including 1) segmenting the trajectory using turning points of all considered variables, and 2) removing segments shorter than the threshold of drivers' reaction time $\tau$. 

\subsection{Time-series Action Phase Probability Modeling}
The length of an \emph{Action phase} can vary and is denoted by the labels ``Long (lg)'' or ``Short (st)'' according to a threshold $\eta$. Consequently, the time label space for \emph{Action phases} is represented as $\mathcal{T} = \{lg, st\}$. Subsequently, \emph{Action phase} can be further described by a two-dimensional label space $S' = \{S, \mathcal{T}\}$, which is taken as input for the Action phase transition probability model (\textbf{Algorithm 3}). Let $S'_{n'}$ and $S'_{{n'}+1}$ represent two adjacent \emph{Action phases}, where $n' \in \{1, 2, \dots, N'-1\}$. The transition probability between them can be mathematically represented as a function $\mathcal{R}\left(S'_{n'}, S'_{{n'}+1}\right)$, which captures the underlying characteristics or patterns between $S'_{n'}$ and $S'_{{n'}+1}$. This transition probability function provides insight into the relationship and dynamics between consecutive \emph{Action phases} in the time-series analysis.

\subsection{Coupled Markov Chain Theory}
Two main approaches are commonly used to implement the transition of driving behaviour segments. The first approach utilises Markov models, including Markov Chains and Hidden Markov Models (HMM) \cite{wang2018driving}, which are easily interpretable and capable of capturing underlying structures. However, when dealing with a large number of hidden layers, HMM may become computationally inefficient and less accurate due to increased complexity. The second approach involves deep learning models such as Recurrent Neural Networks (RNN), Long Short-Term Memory (LSTM), and Gated Recurrent Unit (GRU) Networks \cite{wang2018driving}. These models can address the complexity limitation of HMM and capture complex relationships between \emph{Action phases}. Nevertheless, they typically require a large amount of training data and are computationally expensive due to their gating mechanisms.   

In our case, the Markov Chain method is adopted to implement the Action Probability (\textbf{Algorithm 3}). The concept of a coupled chain refers to the collective behaviour of two independent systems, each following the principles of a classical Markov chain \cite{Billingsley1995}. Let's consider two one-dimensional Markov chains ($X_i$) and ($Y_j$) that operate on the state space $\{S_1, S_2, ..., S_n \}$, with positive transition probabilities defined as 

\begin{equation}\label{eq11}
    \Pr (X_{i+1} = S_k, Y_{j+1} = S_f | X_i = S_l, Y_j = S_m ) = p_{lm, kf}   
\end{equation}
here, the $(X_i)$ chain describes the \emph{Action phase} state $S'$ and the $(Y_i)$ chain describes the time label $\mathcal{T}$. Then the coupled transition probability $p_{lm, kf}$ on the state space $\{S_1, S_2, ..., S_n\} \times \{S_1, S_2, ..., S_n\}$ is given by 

\begin{equation} \label{eq12}
    p_{lm, kf} = p_{lk} \cdot p_{mf}
\end{equation}

Two coupled one-dimensional Markov chains can be utilised to construct a two-dimensional spatial stochastic process on a lattice represented by ($Z_{i,j}$). The lattice consists of a two-dimensional domain of cells, as depicted in Figure \ref{coupled_MC}. The deep blue cells represent known boundary cells, the light blue cells indicate known cells within the domain (past observations), and the white cells represent unknown cells. The future state used to determine the state of cell ($i, j$) is cell ($N_x, j$), where each cell is identified by its row number $i$ and column number $j$. Then the conditional probabilities can be expressed as follows \cite{elfeki2001markov}:

\begin{figure}[!t]
\centering
 \includegraphics[width = 0.8\linewidth]{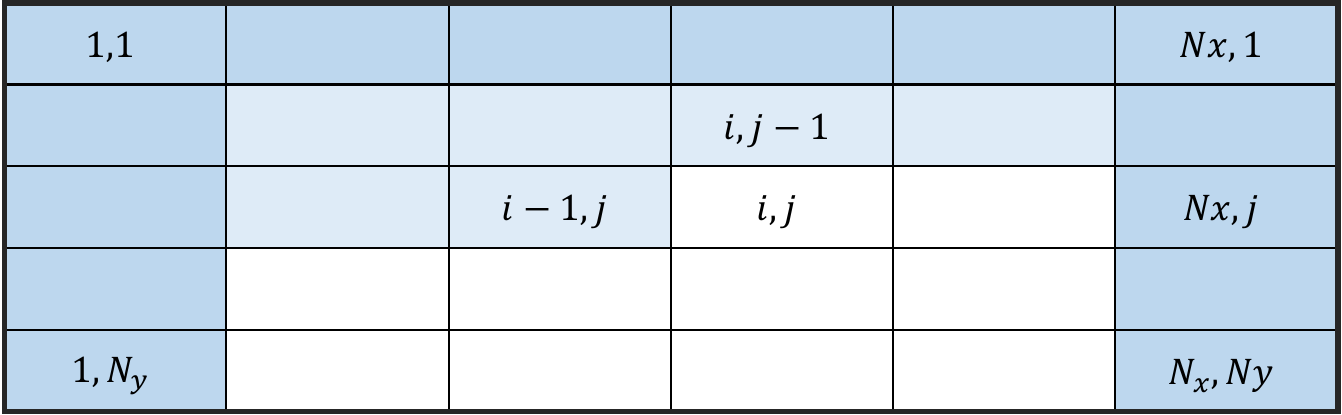}
 \caption{Conditional Markov chain on the states of the future}
 \label{coupled_MC}
\end{figure}

\begin{equation} \label{eq13}
    P_{lk}^{h} = \Pr(X_{i+1} = S_k | X_i = S_l)
\end{equation}

\begin{equation} \label{eq14}
    P_{mk}^{v} = \Pr(Y_{j+1} = S_k | Y_j = S_m)
\end{equation}

The stochastic process $(Z_{ij})$ is obtained by coupling the Markov chains $(X_i)$ and $(Y_j)$ while ensuring that these chains transition to the same states. Therefore, we have:

\begin{equation} \label{eq15}
\begin{aligned}
    & \Pr(Z_{i,j} = S_k | Z_{i-1, j} = S_l, Z_{i, j-1} = S_m) \\
    & = C \, \Pr(X_i = S_k | X_{i-1} = S_l) \, \Pr(Y_{j-1} = S_m)
\end{aligned}
\end{equation}
where $C$ is a normalising constant that arises from restricting transitions in the ($X_i$) and ($Y_j$) chains to the same states. It is calculated as:

\begin{equation} \label{eq16}
    C = \left( \sum_{f = 1}^n p_{lf}^h \cdot p_{mf}^v \right)^{-1}
\end{equation}

By combining Equation \ref{eq15} and Equation \ref{eq16}, the required probability can be expressed as:

\begin{equation} \label{eq17}
\begin{aligned} 
    p_{lm, k} & := \Pr(Z_{i,j}=S_k | Z_{i-1,j} = S_l, Z_{i, j-1}=S_m) \\
    & = \frac{p_{lk}^h \cdot p_{mk}^v}{\sum_f p_{lf}^h \cdot p_{mf}^v}, \, k = 1,...,n
\end{aligned}
\end{equation}

\section{Data-based evaluation}

\subsection{Data Preporcessing}
In this study, the NGSIM highway dataset, which includes data from I-80 and US-101, was utilised to investigate the heterogeneity of longitudinal driving behaviour based on our proposed framework. A comprehensive preprocessing of the dataset, involving filtering and extraction, was conducted as described by Sun et al. \cite{sun2021modeling}. Especially, drivers with trajectories lasting at least 50 seconds were selected to ensure an adequate amount of data for analysing longitudinal driving behaviour \cite{wang2017much}. The final extracted dataset consisted of 123 drivers from the I-80 dataset and 848 drivers from the US-101 dataset.

The driving behaviour variables considered in this study were velocity ($v$), acceleration ($a$), distance ($d$) between the preceding and following vehicles, and their speed difference ($\Delta v$). The threshold values used in \textbf{Algorithm 1} were determined based on empirical knowledge from literature \cite{dingus2006development}, as summarised in Table \ref{threshold}.

\begin{table}[!htb]
  \renewcommand{\arraystretch}{1.3}
  \caption{Parameter settings of \textbf{Algorithm 1}}
  \label{threshold}
  \centering
  \begin{tabular}{|c|c|c|c|c|c|c|}
    \hline
     \textbf{$\Delta y$ /(unit)} & \textbf{$\theta_1$} & \textbf{$\theta_2$} & \textbf{$\delta$} & \textbf{$\gamma$} & \textbf{$\tau$} & \textbf{$\eta$} \\
    \hline
    $v/(m/s)$ & 2 & -2 & 20 & 30 & 10 & 50\\
    \hline
    $a / (m/s^2)$ & 0.25 & -0.25 & 0.25 & 30 & 10 & 50\\
    \hline
    $d / (m)$ & 1 & -1 & 1 & 30 & 10 & 50\\
    \hline
    $\Delta v / (m/s) $ & 2  & -2 & 2 & 30 & 10 & 50\\
    \hline
  \end{tabular}
\end{table}

\subsection{Visualisation and Analysis of Action Phase}

\begin{figure}[!htb]
  \centering
  \subfloat[Driver ID270 from I-80 dataset\label{I80-270}]{
    \includegraphics[width=0.93\columnwidth]{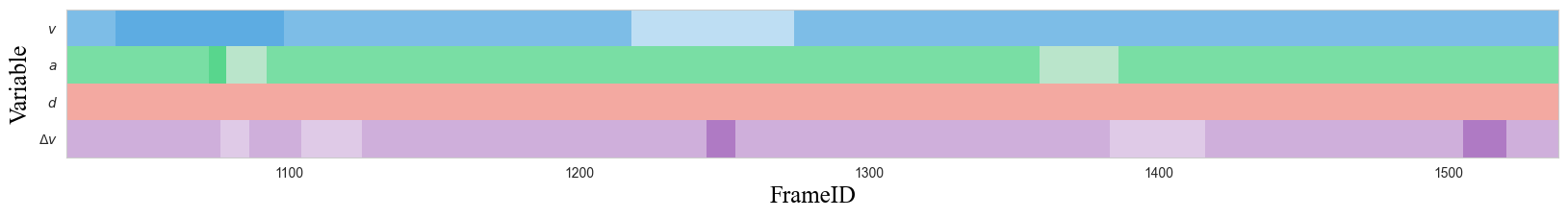}
  }
  % \vspace{1em}
  
  \subfloat[Driver ID1264 from I-80 dataset\label{I80-1264}]{
    \includegraphics[width=0.93\columnwidth]{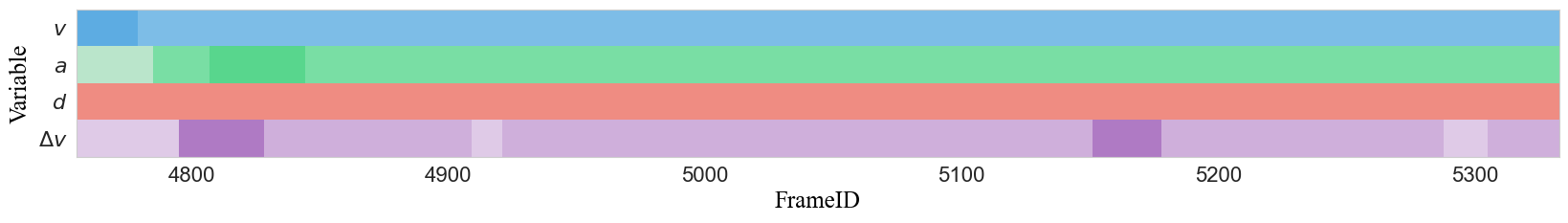}
  }
  % \vspace{1em}
  
  \subfloat[Driver ID33 from US-101 dataset\label{US101-33}]{
    \includegraphics[width=0.93\columnwidth]{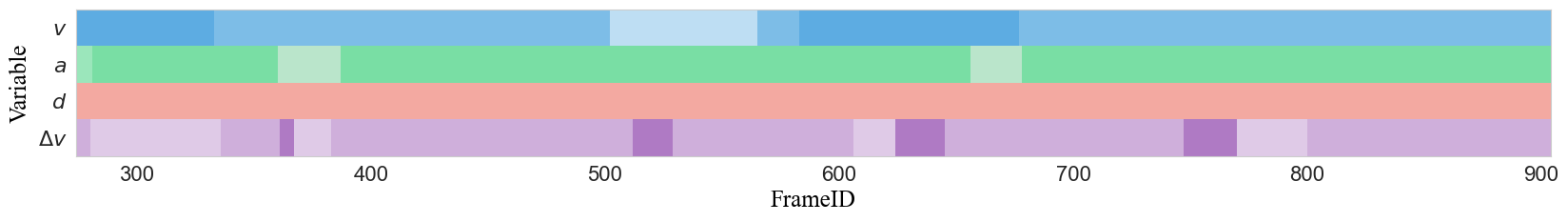}
  }
    % \vspace{1em}
    
  \subfloat[Driver ID1035 from US-101 dataset\label{US101-1035}]{
    \includegraphics[width=0.93\columnwidth]{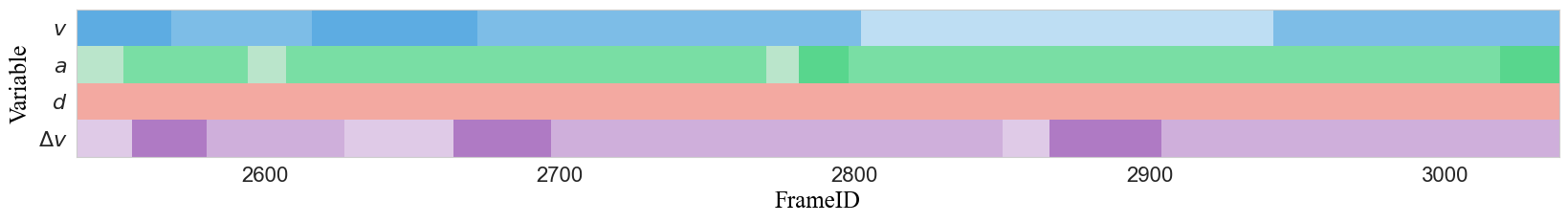}
  }
  \caption{Visualisation of \emph{actions}: the driving behaviour map}
  \label{drivingmap}
\end{figure}

The \emph{action trend} labels for the four driving behaviour variables are obtained using \textbf{Algorithm 1}. These results are then visualised, generating unique driving behaviour maps for each driver, as exemplified in Figure \ref{drivingmap}. In the figure, various colours represent different driving behaviour variables, with velocity, acceleration, distance, and speed difference represented in that order. The varying intensity of the same colour indicates different \emph{action trend} names, including Increasing, stable as High, stable as Low, and Decreasing.

In Figure \ref{I80-270}, the dominant \emph{action trend} for acceleration is ``L'', although instances of ``I'' and ``D'' can also be observed. The distance remains relatively stable without frequent \emph{action trend} changes. When comparing the driving behaviour maps of the four drivers shown, driver ID1264 from the I-80 dataset exhibits the fewest \emph{action trend} changes across the four variables. Conversely, drivers ID3 and ID1035 from the US-101 dataset demonstrate a higher frequency of the changes.

The driving behaviour map offers an intuitive approach to interpreting driving behaviour by visualising the changes in driving behaviour over time. It is also important to note that this visualisation method relies on observation and should be complemented with further quantitative evaluation, which will be carried out in subsequent steps.

The \emph{Action phase} Library for a specific traffic flow was constructed using \textbf{Algorithm 2}. The resulting Library consists of 142 \emph{Action phases} for the I-80 dataset and 228 \emph{Action phases} for the US-101 dataset. Table \ref{action-results} presents the top 10 \emph{Action phase} along with their corresponding frequencies. Notably, both traffic flows exhibit a significant overlap in their high-frequency \emph{Action phases}, and the top three \emph{Action phases} are identical for both datasets. These top \emph{Action phases} include ``((L, L, H, H), st)'', ``((L, L, H, H), lg)'', and ``((L, L, L, H), st)'', which indicate common driving behaviour across the datasets. The reason behind this is that the two datasets were collected during evening and morning rush hours respectively. In these periods, the density of traffic flow is significantly high, with most vehicles exhibiting car-following behaviour and even close to congestion. Due to this, there is limited variability in driving behaviours, resulting in a scarcity of ``I'' and ``D'' and a high frequency of ``Stable (H and L)''. The high-density traffic flows also provide an explanation for the highest frequency of occurring ``L''. 

\begin{table}[!htb]
    \centering
    \caption{Statistics of \emph{Action phase} (Top 10)}
    \label{action-results}
    \begin{tabular}{|c|c|c|c|}
        \hline
        \multicolumn{2}{|c|}{\textbf{I-80}} & \multicolumn{2}{c|}{\textbf{US-101}} \\ \hline
        \emph{\textbf{Action phase}} & \textbf{Frequency} & \emph{\textbf{Action phase}} & \textbf{Frequency} \\ \hline
        ((L,L,H,H), st) & 415 & ((L,L,H,H), st) & 2703 \\ \hline
        ((L,L,H,H), lg) & 219 & ((L,L,H,H), lg) & 1661 \\ \hline
        ((L,L,L,H), st) & 156 & ((L,L,L,H), st) & 965 \\ \hline
        ((L,L,H,I), st) & 68 & ((L,L,L,H), lg) & 672 \\ \hline
        ((L,L,L,H), lg) & 65 & ((D,L,H,H), st) & 651 \\ \hline
        ((D,L,H,H), st) & 54 & ((L,L,L,L), st) & 480 \\ \hline
        ((L,L,L,L), st) & 41 & ((I,L,H,H), st) & 469 \\ \hline
        ((L,L,H,D), st) & 39 & ((D,L,H,H), lg) & 419 \\ \hline
        ((D,L,H,I), st) & 38 & ((D,L,H,I), st) & 412 \\ \hline
        ((D,I,H,I), st) & 30 & ((L,L,H,I), st) & 349 \\ \hline
    \end{tabular}
\end{table}

\subsection{Analysis of Action-chain}
The transition probabilities from one \emph{Action phase} to another within the \emph{Action phase} Library were computed using \textbf{Algorithm 3}. Some \emph{Action phases} either have no transitions or exhibit very low probabilities of transition. Conversely, other \emph{Action phases} tend to be transitioned to by a greater number of \emph{Action phases}. The results adhere to the fundamental principles of driving behaviour. For example, the \emph{Action phase} ``((L, L, L, H), st)'' from the US-101 dataset demonstrates higher probabilities of being transitioned. This can be attributed to the fact that the driving data were collected during the morning peak hour when there is typically high traffic flow density, leading drivers to adopt more consistent driving behaviours with lower values.

Overall, each \emph{Action phase} was found to have a following \emph{Action phase} with the highest transition probability, resulting in the formation of an \emph{Action-chain}, as illustrated in the examples provided in Table \ref{action-porb}. In the I-80 dataset, for instance, the \emph{Action phase} ``((D, I, I, H), st)'' has a probability of 0.68 to transition to ``((L, I, I, H), st)'', which is higher than any other \emph{Action phases}.

\begin{table}[!htb]
    \centering
    \caption{\emph{Action-chain} composed by the highest joint transition probability (JTP)}
    \label{action-porb}
    \begin{tabular}{|c|c|c|c|}
        \hline
        \textbf{Dataset} & \textbf{\emph{Action phase} from} & \textbf{\emph{Action phase} to} & \textbf{JTP} \\ \hline
        \multirow{5}{*}{I-80} & ((D, D, I, I), st) & ((D, L, L, I), st) & 0.68 \\ \cline{2-4}
        & ((D, I, I, H), st) & ((L, I, I, H), st) & 0.68 \\ \cline{2-4}
        & ((D, I, D, I), lg) & ((L, L, D, H), st) & 0.67 \\ \cline{2-4}
        & \ldots & \ldots & \ldots \\ \cline{2-4}
        & ((D, D, H, D), lg) & ((L, L, H, H), st) & 0.52 \\ \hline
        \multirow{5}{*}{US-101} & ((D, D, D, I), st) & ((L, L, L, L), st) & 0.64 \\ \cline{2-4}
        & ((D, H, L, H), st) & ((L, L, L, H), st) & 0.64 \\ \cline{2-4}
        & ((I, I, L, D), st) & ((L, L, L, H), st) & 0.58 \\ \cline{2-4}
        & \ldots & \ldots & \ldots  \\ \cline{2-4}
        & ((I, I, D, I), st) & ((L, L, D, L), st) & 0.32 \\ \hline
    \end{tabular}
\end{table}

\subsection{Evaluation of Driving Heterogeneity}

In this study, we assume that the maximum transition probability represents the generally adopted \emph{Action phase} of drivers in a specific traffic flow, indicating the average level of driving behaviour. However, in the real world, drivers often deviate from this general level due to heterogeneity in their driving behaviours. To quantify the heterogeneity, we define this heterogeneity as the deviation between the actual \emph{Action phase} transition and the \emph{Action-chain}. 

The Mean Squared Error (MSE) is a commonly used method for measuring the average squared difference between two sets of data, and it serves as the metric to quantify driving heterogeneity in this context, see Equation \ref{mse}. 

\begin{equation} \label{mse}
    DH = \frac{1}{N'} \times \sum^{N'}_{n'=1} (P'_{n'} - P_{n'})^2
\end{equation}
where $N'$ is the total number of \emph{Action phases}, and $P'_{n'}$ and $P_{n'}$ represent the transition probability of actual \emph{Action phase} and the maximum transition probability, respectively. A higher value indicates a greater driving heterogeneity. 

\subsection{Numerical Results and Discussions}
\begin{figure}[!htb]
  \centering
  \subfloat[I-80 dataset\label{I80-hetero}]{
    \includegraphics[width=0.44\columnwidth]{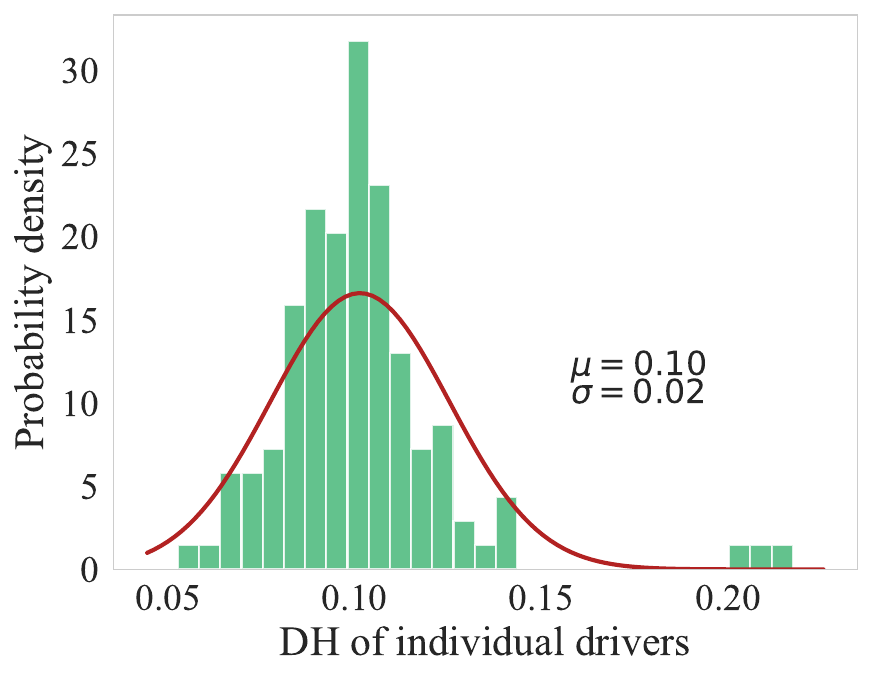}
  }
  \subfloat[US-101 dataset\label{US101-hetero}]{
    \includegraphics[width=0.44\columnwidth]{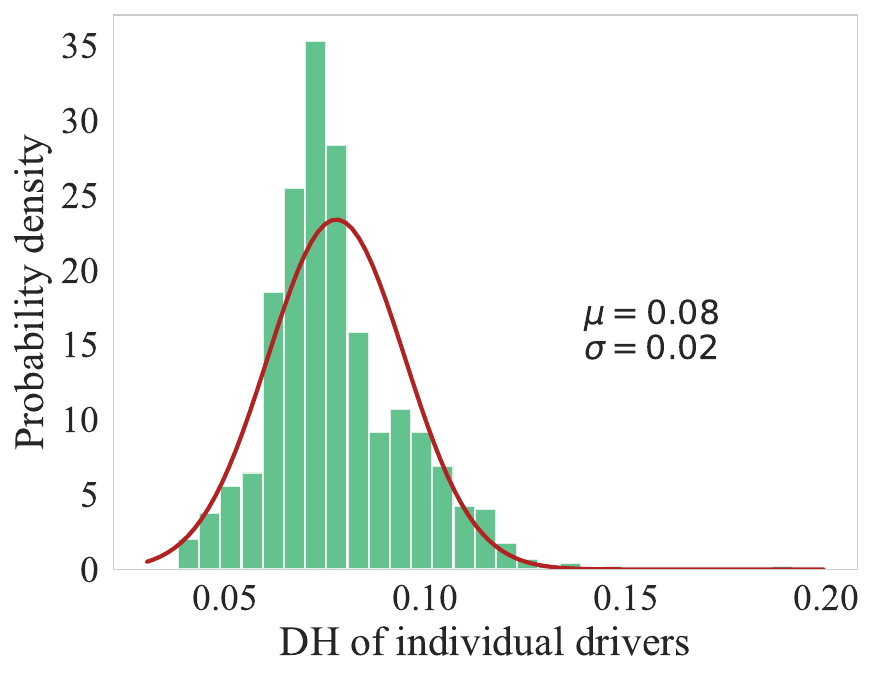}
  }
  \caption{Driving heterogeneity in a specific traffic flow}
  \label{hetero}
\end{figure}

The heterogeneity of individual drivers in a specific traffic flow was calculated and subjected to further statistical analysis using the normal distribution. The $3\sigma$ rule of thumb is commonly employed in data analysis to identify potential outliers or unusual behaviour. By applying the $3\sigma$ principle, drivers with atypical driving behaviour were identified, as summarised in Table \ref{3-sigma}. These drivers may serve as potential factors contributing to increased traffic flow heterogeneity and negatively affecting traffic performance. It is also noteworthy that the degree of driving heterogeneity in the two traffic flows exhibits the same variance, see Figure \ref{hetero}; however, the drivers on US-101 (with $\mu = 0.08$) display overall lower levels of heterogeneity compared to those on I-80 (with $\mu = 0.10$). 

\begin{table}[!htb]
    \centering
    \caption{Drivers with the highest heterogeneity}
    \label{3-sigma}
    \begin{tabular}{|c|c|c|c|}
        \hline
        \multicolumn{2}{|c|}{\textbf{I-80}} & \multicolumn{2}{c|}{\textbf{US-101}} \\ \hline
        \textbf{Driver ID} & \textbf{$DH$} & \textbf{Driver ID} & \textbf{$DH$} \\ \hline
        295 & 0.2172 & 582 & 0.1379 \\ \hline
        535 & 0.2111 &628 & 0.1360 \\ \hline
        1174 & 0.2004 & 1157 & 0.1464 \\ \hline
        - & - & 1647 & 0.1916 \\ \hline
    \end{tabular}
\end{table}

\section{Conclusion}

In this study, a novel framework was proposed to address driving heterogeneity in a comprehensive manner. By introducing the concepts of \emph{Action phase} and \emph{Action-chain}, along with specialised algorithms, the framework effectively quantified and explained driving heterogeneity at both the individual driver and traffic flow levels. Real-world datasets were used for evaluation, validating the framework's ability to offer clear interpretations. Although the contributions of novel insights and findings of heterogeneity of driving behaviour, further validation and justification of the methods employed in each step are still required, which is a focus of our ongoing research.

% Generated by IEEEtran.bst, version: 1.14 (2015/08/26)

\bibliographystyle{IEEEtran}
% \bibliography{reference}

\end{document}